%% file: main.tex
\documentclass[letterpaper, 10 pt, conference]{ieeeconf}  

\IEEEoverridecommandlockouts                              
\usepackage{graphicx}
\usepackage{amsmath}
\usepackage{amssymb}
\usepackage{amsfonts}
\usepackage{amsthm}
\usepackage{subcaption}

\usepackage{algorithm}
\usepackage{algorithmic}

\usepackage{xcolor}
\usepackage[capitalize]{cleveref}
\usepackage{cite}

\newtheorem{theorem}{Theorem}[section]

\newtheorem{definition}[theorem]{Definition}

\crefname{section}{Section}{Sections}
\crefname{theorem}{Theorem}{Theorems}
\crefname{lemma}{Lemma}{Lemmas}
\crefname{table}{Table}{Tables}
\crefname{equation}{Eq}{Eqs}
\crefname{algocf}{Algorithm}{Algorithms}
\Crefname{algocf}{Algorithm}{Algorithms}
\crefname{ALC@unique}{Line}{Lines}

\title{\LARGE \bf
Verification of Visual Controllers\\via Compositional Geometric Transformations
}

\author{Alexander Estornell, Leonard Jung, Michael Everett
\thanks{Authors with Northeastern University, Boston, MA, USA. e-mail: {\tt\footnotesize \{estornell.a,\allowbreak jung.le,\allowbreak m.everett\}\allowbreak@\allowbreak northeastern.edu}.
This work was supported by award W911NF-24-2-006.
}%
}

\begin{document}

\maketitle

\begin{abstract}
Perception-based neural network controllers are increasingly used in autonomous systems that rely on visual inputs to operate in the real world. 
Ensuring the safety of such systems under uncertainty is challenging.
Existing verification techniques typically focus on $L^p$-bounded perturbations in the pixel space, which fails to capture the low-dimensional structure of many real-world effects.
In this work, we introduce a novel verification framework for perception-based controllers that can generate outer-approximations of reachable sets through explicitly modeling uncertain observations with geometric perturbations. Our approach constructs a boundable mapping from states to images, enabling the use of state-based verification tools while accounting for uncertainty in perception. 
We provide theoretical guarantees on the soundness of our method and demonstrate its effectiveness across benchmark control environments. 
This work provides a principled framework for certifying the safety of perception-driven control systems under realistic visual perturbations.
\end{abstract}

\section{Introduction}
Visual inputs, or other high-dimension perception, are commonly used to infer states and make decisions in modern autonomous systems, ranging from self-driving cars and warehouse robots to aerial drones.
More recently, there has been growing interest in policies that map directly from perception data to control actions (``perception-to-action'' or ``pixels-to-torques'' controllers).
This new control paradigm has major upside, as it can substantially reduce the computational and design effort of more traditional, modular architectures.
These policies may be trained on human demonstrations, to imitate slower modular architectures, or with reinforcement learning.
While expensive to train, deploying the policy simply requires a forward pass when new perception data arrives, which can be lightweight enough for resource-constrained control systems and can reduce latency for extremely agile control.

However, a core difficulty lies in formally analyzing the behavior of these controllers in response to uncertainty or changes in their visual inputs.
Existing techniques to verify properties of neural networks are predominantly designed for robustness against $L^p$ norm-bounded perturbations to the input \cite{zhang2018efficient, weng2018towards, xu2020automatic, raghunathan2018semidefinite, tjeng2017evaluating, katz2019marabou, katz2017reluplex, jia2021verifying,vincent2021reachable,DeepPoly19,IBP19}. 
While theoretically convenient, $L^p$ perturbations (e.g., imperceptible pixel-level changes) are poorly aligned with the kinds of real-world uncertainty that perception systems encounter. 
For example, slight variation in camera angle could cause every pixel in the image to change; $L^p$-based bounds in the pixel space would need to be vacuously loose to capture this scenario.

Instead, this paper investigates uncertainty through the lens of geometric transformations (e.g., translation, rotation, scaling), which more naturally capture the types of movements that objects will undergo in the real world. 
These geometric transformations enable the mapping of low dimensional uncertainty (e.g., uncertainty in the angle of an arm) to high dimensional observations, rather than relying on the articulation of pixel-wise uncertainty, which is typically impossible without $L^P$-assumptions.
As such, verification under these transformations more accurately reflect the dynamics of physical environments, such as movement of a camera, objects moving in a scene, or uncertain pose measurements.

A key issue is that real-world scenes often involve object-specific variations. That is, different objects within a scene may be subject to distinct transformations—for example, one object may rotate while another is translated.
Much of the recent work on this topic is limited to geometric transformations across the whole image~\cite{DeepG19, Semantify20, li2021tss}.
Meanwhile, other works learn to approximate complicated image transformations, but it is difficult to know how accurate those new mappings are \cite{GAN22, sevin25}.
To rigorously model such scenarios, we propose to apply geometric transformations at the object (or entity) level, rather than uniformly across the entire image. 
Our approach considers each entity separately and applies appropriate transformations to each, then analyzes the decomposed scene and ultimately recomposes the results into a single consistent output.

Extending this analysis from a single uncertain image to a learned control policy over a time horizon presents further challenges.
For example, given an initial uncertainty set in the system's state, the verifier needs to reason about the entire set of possible image observations the controller may receive at the current and future timesteps.
Existing work on verifying learned control policies is mostly limited to state-based controllers (i.e., mapping low-dimensional state to action), which avoid this issue but cannot leverage high-dimensional perceptual inputs.

To address these challenges, this paper proposes a new approach consisting of two primary stages. First, we construct pixel-wise linear bounds on the input image to the controller, derived from entity-specific geometric transformations applied to a decomposed scene. Each entity in the environment (e.g., links in Acrobot) is individually bounded under its own parameterized transformation, which provides linear bounds on both the RGB and $\alpha$ channels. These per-entity bounds are then composed using alpha compositing to yield pixel-wise linear bounds on the final blended observation image. Second, we propagate these image bounds through the controller and system dynamics. This enables computing bounds on the control action and subsequently on the future system state. By composing both stages, we perform closed-loop reachability analysis, producing a true outer approximation of the set of future states reachable under initial uncertainty and geometric perceptual variation.

In summary, our contributions include:
\begin{enumerate}
    \item \textbf{A compositional verification framework for perception-based controllers.}  
    We introduce a novel pipeline that constructs pixel-wise linear bounds over perception inputs by decomposing the scene into semantically meaningful entities, applying entity-specific geometric transformations, and recomposing the full observation via alpha-blending.

    \item \textbf{Integration of image-based perturbation bounds with state-space reachability analysis.}  
    By bounding the image space under geometric transformations and composing these bounds with learned perception and dynamics models, we extend formal state-based verification methods to operate in vision-based closed-loop systems.

    \item \textbf{Soundness guarantees and compatibility with existing neural network verifiers.}  
    We provide formal guarantees that our method yields outer approximations of reachable states.

    \item \textbf{Empirical validation on benchmark control tasks.}  
    We demonstrate our approach on standard Gym environments (CartPole, Pendulum, Acrobot), to highlight the scalability with image size and tight reachable set estimates under perceptual uncertainty. 
\end{enumerate}

\section{Preliminaries}
\input{preliminaries}

\section{Methodology}\label{sec:method}

\begin{figure*}
    \centering
    \includegraphics[width=0.8\textwidth]{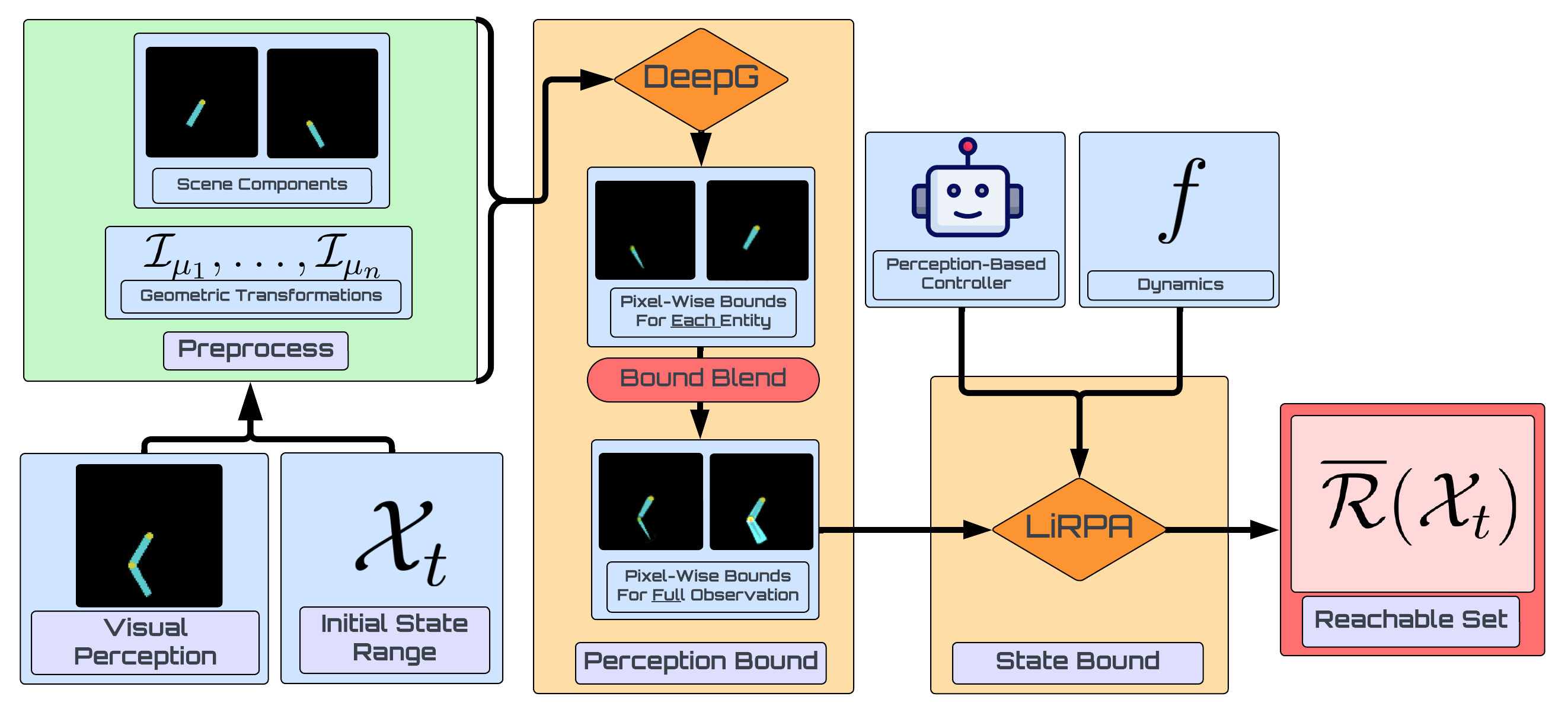}
    \caption{Verification Pipeline. Given an initial state range, the proposed algorithm calculates bounds on future states.}
    \label{fig:method}
\end{figure*}

We now describe our pipeline for verifying perception-based controllers by computing outer bounds on reachable sets under geometric transformations. An overview is provided in ~\cref{alg:verify} and illustrated in \cref{fig:method}.
Recall that for a control policy $\pi_{\theta}$, system dynamics $f$, and observation process $o$, our goal is to compute an outer bound $\bar{\mathcal{R}}_T(\mathcal{X}_t)$ on the set of states reachable within $T$ timesteps, starting from a set of initial possible states $\mathcal{X}_t$ (e.g., arm angle between 40$^\circ$ and 45$^\circ$).
We focus on settings where the observation $\mathbf{y}_t = o(\mathbf{x}_t)$ is an image.\footnote{In practice, observations may include both visual and low-dimensional state inputs (e.g., velocity). 
These components can be incorporated by concatenation, treating them as additional pixels.}

To construct the reachable set bound $\bar{\mathcal{R}}_T(\mathcal{X}_t)$, each component in the feedback loop (\cref{eq:neural_feedback_loop})
must be bounded. 
We build upon existing methods to bound $f$ and $\pi_{\theta}$, and focus on 1) bounding the observation process using geometric transformations, and 2) composing bounds on $f$, $\pi_\theta$, and $o$ to produce the final reachable set bound $\bar{\mathcal{R}}_T(\mathcal{X}_t)$.

\subsection{Geometric Transformations}

To bound the observation process $o$, we model visual scenes as composed of individual entities (e.g., object parts), each governed by simple geometric motions.
Within an image $\mathbf{y}$, these motions are described by geometric transformations of the form $\mathcal{I}_{\psi, \varphi, \mu}$, where $\mu$ is a spatial transformation parameter 
(e.g., a rotation angle or translation offset) and $\psi$, $\varphi$ are parameters for brightness and contrast. In this work, we consider only spatial transformations without loss of generality.
Applying the transformation to a pixel $\mathbf{y}[l, k]$ yields a new pixel value defined by:
\begin{equation}
\mathcal{I}_{\mu_i}\big(\mathbf{y}[l, k]\big) := \mathbf{y}'[l, k] \nonumber
\end{equation}
To model that each object can move individually (rather than applying one transformation to the entire image), we decompose the scene into regions such that several transformations can be applied independently to each entity.

Analyzing each entity independently is achieved through an \emph{environmental initialization} step\footnote{This initialization only needs to be performed once per environment and can then be reused for any controller $\pi_\theta$.}, in which we assign entity $i$ a mask $\alpha_i$ identifying the associated pixels, along with a corresponding transformation $\mathcal{I}_{\mu_i(\mathbf{x})}$ where $\mu_i$ is a mapping from states $\mathbf{x}$ to the transformation parameters (e.g., a mapping from arm angle $\theta$ to the angle of rotation of the image). 
The transformation $\mathcal{I}_{\mu_i(\mathbf{x})}$ governs the way in which the perception of entity $i$ changes as a function of underlying state $\mathbf{x}$.
Hence, the set of transformation parameters $ \boldsymbol{\mu}(\mathbf{x}) = \langle \mu_1(\mathbf{x}), \ldots \mu_n(\mathbf{x})\rangle$ is a per-entity parametrization of the full observation process $o(\mathbf{x})$, that is, we can write
\begin{align}
    o(\mathbf{x}) = O\big(\mathcal{I}_{\mu_1(\mathbf{x})}, \ldots, \mathcal{I}_{\mu_n(\mathbf{x})} \big),
\end{align}
for some function $O$ which maps from the space of geometrically transformed entities to the set of observations.

Since these geometric transformations are parameterized by the state $\mathbf{x}$, we can express the range of transformation parameters for a given initial state range $\mathcal{X}_t$ as the set 
\begin{align}
\boldsymbol{\mu}(\mathcal{X}_t) := \big\{\langle \mu_1(\mathbf{x}), \ldots \mu_n(\mathbf{x})\rangle: \mathbf{x} \in \mathcal{X}_t\big\}\label{eq:param}
\end{align}
gives the \emph{initial latent state rage}.
Since $\boldsymbol{\mu}(\mathcal{X}_t)$ is low dimensional, this set can be approximated using standard bound propagation techniques.
For ease of notation we write $\mathcal{I}_{\mu_i(\mathbf{x}_t)}$ simply as $\mathcal{I}_{\mu_i}$.

\begin{figure}[t]
\centering
\begin{subfigure}{0.3\linewidth}
    \centering
    \includegraphics[height=2.5cm]{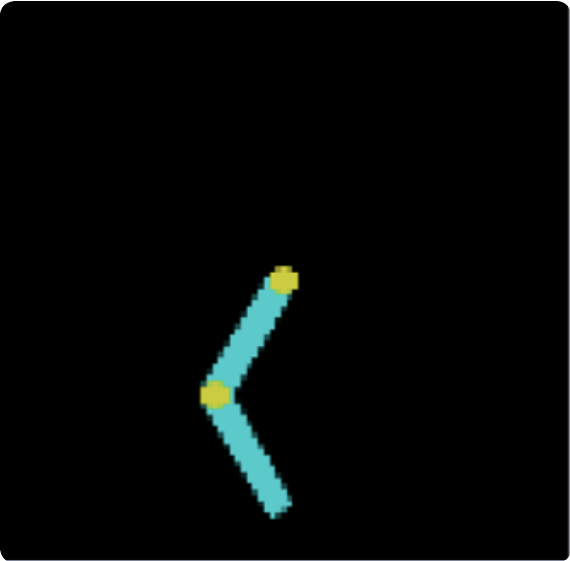}
    \caption{Full Scene}
    \label{fig:decomp:full_scene}
\end{subfigure}%
\begin{subfigure}{0.7\linewidth}
    \centering
    \includegraphics[height=2.5cm]{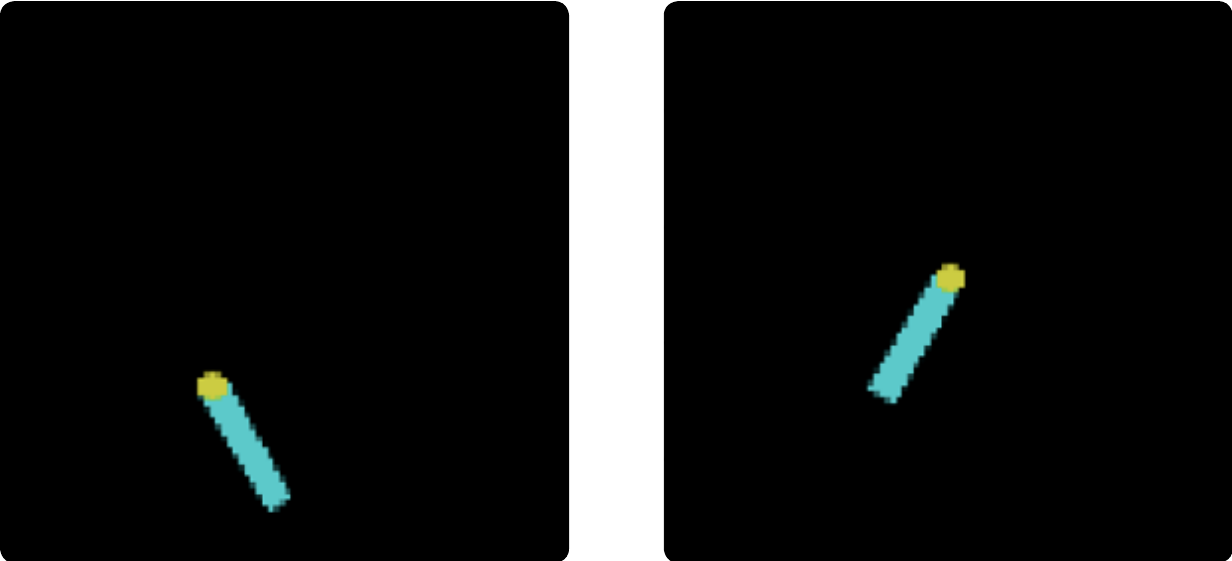}
    \caption{Decomposed Scene}
    \label{fig:decomp:entities}
\end{subfigure}%
\caption{An environment decomposed into entities. This paper proposes entity-specific masks (e.g., for each arm) as a way to efficiently transfer low-dimensional uncertainty (in the state space) to the high-dimensional observation space.}
\label{fig:decomp}
\end{figure}

\textbf{Example: Acrobot.}\emph{
Consider the Acrobot system, which consists of two linked arms. 
Let arm 1 (with joint angle $\theta_1$ and pixel length $r$) be entity 1, and arm 2 (with joint angle $\theta_2$) be entity 2.
We define masks $\alpha_1$ and $\alpha_2$ for the corresponding pixels in the image (see \cref{fig:decomp}).
\begin{itemize}
\item For arm 1, the transformation $\mathcal{I}_{\mu_1}$ encodes rotation about the origin, with $\mu_1 = \theta_1$.
\item For arm 2, the transformation $\mathcal{I}_{\mu_2}$ captures both the rotation $\theta_2$ and translation induced by arm 1. 
Specifically, $\mu_2 = \langle r \cos(\theta_1), r \sin(\theta_1), \theta_2 \rangle$ encodes a rotation by $\theta_2$ followed by a translation by the endpoint of arm 1.
\end{itemize}
While Acrobot's state contains four variables $\theta_1, \theta_2$ (angle) and $\dot{\theta}_1, \dot{\theta}_2$ (angular velocity), the initial latent state range depends only on $\theta_1, \theta_2$,. Thus, if the initial state range $\mathcal{X}_t$ has $45^\circ \leq \theta_1 \leq 50^\circ$ and $10^\circ \leq \theta_2\leq 13^\circ$, then we write
\begin{align}
\boldsymbol{\mu}(\mathcal{X}_t) := \big\{\langle \mu_1(\theta_1, \theta_2),~\mu_2(\theta_1, \theta_2)\rangle:\, &\theta_1 \in [45^\circ, 50^\circ],\nonumber\\&\theta_2 \in[ 10^\circ, 13^{\circ} ] \big\}.\nonumber
\end{align}
}

In general, for an environment with $n$ entities, we define a set of masks $\alpha_1, \ldots, \alpha_n$ and associated transformations $\mathcal{I}_{\mu_1}, \ldots, \mathcal{I}_{\mu_n}$ that describe how state changes affect the image $Y$ (e.g., arm rotation can be treated as image rotation). 
Thus, after applying each geometric transformation to its entity, we can \emph{recompose} the individual entities into a full scene representing the observation process,
\begin{align}
  o(\mathbf{x}) & = O\big(\mathcal{I}_{\mu_1(\mathbf{x})}, \ldots, \mathcal{I}_{\mu_n(\mathbf{x})} \big) = Y \label{eq:obs_param}\\
&= Y_{0} \cdot \prod_{i=1}^{n} (1 - \mathcal{I}_{\mu_i}(\alpha_i)) \label{eq:alpha_blend}\\
&\qquad + \sum_{i=1}^{n} \left( \mathcal{I}_{\mu_i}(\alpha_i) \cdot \mathcal{I}_{\mu_i}(Y_i) \cdot \prod_{j=i+1}^{n} (1 - \mathcal{I}_{\mu_j}(\alpha_j)) \right)\nonumber
\end{align}
With this formulation of the observation $Y$, our goal will thus be to first attain pixel-wise bounds on the observation process, which can then be propagated through the controller and dynamics, finally obtaining bounds on future states.

\subsection{Pixel-Wise Bounds via Geometric Transformations}
Our ultimate goal is to provide bounds on future reachable sets.
Before that, we need to understand how the control policy $\pi_{\theta}$ will act under a range of possible observations. 
We utilize \cref{eq:alpha_blend} to compute affine bounds on pixels over a range of behaviors induced by the geometric transformations, e.g., to bound a given pixel $\mathbf{y}[l, k]$ under a rotation transformation $\mathcal{I}_{\mu}$ where $\mu \in [40^{\circ}, 45^{\circ}]$.

As mentioned previously, bounding geometric transformations using pixel-wise bound propagation techniques (such as CROWN) can require intractable memory and produce vacuously loose bounds. 
Thus, rather than directly applying such a technique to \cref{eq:alpha_blend}, we propose a hybrid approach in which we first leverage a more focused verifier such as DeepG \cite{DeepG19} (which is capable of efficiently producing close to optimal bounds for pixels under geometric transformations) 
to transform \cref{eq:alpha_blend} such that it can be easily handled by existing bound propagation techniques.
In particular, we use DeepG to transform \cref{eq:alpha_blend} into a set of bounds which are then able to be integrated as intermediate results for CROWN. 
We next outline how we attain and use these bounds.

For each of the $n$ entities in the image, we apply DeepG to entity $Y_i$ with mask $\alpha_i$ acted by geometric transformation $\mathcal{I}_{\mu_i}$, producing linear bounds on the pixels $Y[l, k]$, namely $[\underline{Y}_{i, l, k}, \overline{Y}_{i, l, k}]$, and the mask $[\underline{\alpha}_{i, l, k}, \overline{\alpha}_{i, l, k}]$, where 
\begin{align}
\underline{Y}_{i, l, k} &= w_{i, l, k}^{\ell} \mu_i + b_{i, l, k}^{\ell}&, \quad \underline{\alpha}_{i, l, k} &= w_{i, l, k}^{\ell} \mu_i + b_{i, l, k}^{\ell}\nonumber\\
\overline{Y}_{i, l, k} &= w_{i, l, k}^{\text{u}} \mu_i + b_{i, l, k}^{\text{u}}&,\quad
\overline{\alpha}_{i, l, k} &= w_{i, l, k}^{\text{u}} \mu_i + b_{i, l, k}^{\text{u}}\label{eq:pixel_wise_bounds}
\end{align}
to get pixel-wise affine bounds for each of the $n$ entities.

To obtain bounds on future states, we then need to blend these pixel-wise affine bounds for each entity $[\underline{Y}_{i}, \overline{Y}_{i}]$, and $[\underline{\alpha}_{i}, \overline{\alpha}_{i}]$, into a single set of pixel-wise bounds for the full observation. 
To do this, we leverage CROWN to bound the bilinearities in \cref{eq:alpha_blend} and use the affine bounds as relaxations over the pixel values and alpha channels; this results in affine bounds for \cref{eq:alpha_blend}, i.e., bounds on the full observation process $o(\mathbf{x}_t)$, namely $[\underline{Y}, \overline{Y}]$.

As outlined in \cref{sec: nn-ver}, the standard CROWN algorithm linearly relaxes a given function $g:\mathcal{B}\rightarrow \mathcal{C}$ over a specified input range $B\subset\mathcal{B}$, producing \textit{linear bounds} over $g(B)$, which are then concretized to produce a \textit{range} over $g(B)$. 
We leverage our modified version of CROWN that instead bounds functions given pre-computed linear bounds. In particular, for the function $g\circ h:\mathcal{A}\rightarrow \mathcal{C}$, where $h: \mathcal{A} \rightarrow \mathcal{B}$, our modified CROWN can bound $g(h(A))$, when given pre-computed linear bounds on $h$, say $[\underline{h}, \overline{h}]$ (which are defined over set $A\subset \mathcal{A}$).

This enables using the bounds provided by DeepG computing reachable sets. In our context, the function $g\circ h$ is the parametrized observation process given in \cref{eq:obs_param}, i.e., 
\[g\circ h = O\circ (\mathcal{I}_{\mu_1},\,\ldots,\,\mathcal{I}_{\mu_n}),
\]
and the pre-computed linear bounds from DeepG  \cref{eq:pixel_wise_bounds},
\[
[\underline{h}, \overline{h}] = \big\langle [\underline{Y}_1, \underline{\alpha}_1, \overline{Y}_1, \overline{\alpha}_1], \ldots, [\underline{Y}_n, \underline{\alpha}_n, \overline{Y}_n, \overline{\alpha}_n]\big\rangle,
\]
which allows us to linearly relax the function $O$ over the precomputed linear bounds' input range, $\boldsymbol{\mu}(\mathcal{X}_t)$. 
We use this custom implementation to bound the observation process over the specified range of latent states with each pre-computed pixel bound as input to the blending process, \cref{eq:alpha_blend}.

\subsection{Bounds on Future States}

In the second stage of our pipeline, we compute bounds on the next reachable state using the affine pixel-wise bounds produced by the first stage $[\underline{Y}, \overline{Y}]$, along with the controller $\pi_{\theta}$ and dynamics model $f$.
We again deploy our modified version of CROWN for bounding the composition of function $g\circ h$ when given pre-computed linear bounds on $h$.
In this case we now have $g \circ h = \pi_{\theta} \circ o$, with pre-computed linear bounds on the true observation process $o$ from our previous step, namely $[\underline{h}, \overline{h}] = [\underline{Y}, \overline{Y}]$. This yields a bounds on control outputs:
\[
[\underline{u}, \overline{u}] \supseteq \pi_{\theta}([\underline{Y}, \overline{Y}]).
\]
The resulting control bounds can then be combined with the initial state range $\mathcal{X}_t$ and passed through the system dynamics model $f$. This yields bounds on the next state, which can be concretized for the one-step reachable set:
\[
\bar{\mathcal{R}}_1(\mathcal{X}_t) \supseteq f(\mathcal{X}_t, [\underline{u}, \overline{u}]).
\]

\begin{algorithm}[tbh]
\caption{Verification of Vision-Based Control Policies under Geometric Uncertainties}
\label{alg:verify}
\begin{algorithmic}[1]
\REQUIRE Initial state range $\mathcal{X}$, controller $\pi$, dynamics $f$, scene entities $\{\mathcal{I}_i\}_0^k$, geometric transformation set $\{\mathcal{I}_i\}_0^k$
\STATE $[\underline{\mathbf{x}}_{\text{lat}}, \overline{\mathbf{x}}_{\text{lat}}] \leftarrow \texttt{getLatentStateRange}\left([\underline{\mathbf{x}}_0, \overline{\mathbf{x}}_0]\right)$
\FOR{$i$ in $[k]$}
\STATE $[\underline{Y}_i, \overline{Y}_i], [\underline{\alpha}_i, \overline{\alpha}_i] \leftarrow \texttt{getPxlBounds}\left(\mathcal{I}_i, [\underline{\mathbf{x}}_{\text{lat}}, \overline{\mathbf{x}}_{\text{lat}}]\right)$
\STATE $[\underline{Y}, \overline{Y}] \leftarrow \texttt{relaxBlend}([\underline{Y}, \overline{Y}], [\underline{Y}_i, \overline{Y}_i], [\underline{\alpha}_i, \overline{\alpha}_i])$
\ENDFOR
\STATE $[\underline{u}, \overline{u}] \leftarrow \texttt{relaxControl}\left(\pi,[\underline{Y}, \overline{Y}]\right)$
\STATE $[\underline{\mathbf{x}}', \overline{\mathbf{x}}'] \leftarrow \texttt{propagateDynamics}\left(f, [\underline{\mathbf{x}}_0, \overline{\mathbf{x}}_0], [\underline{u}, \overline{u}]\right)$
\RETURN 1-step reachable set bounds, $[\underline{\mathbf{x}}', \overline{\mathbf{x}}']$
\end{algorithmic}
\end{algorithm}
\begin{figure*}[h]
\centering
\begin{subfigure}{0.33\linewidth}
    \centering
    \includegraphics[width=\linewidth]{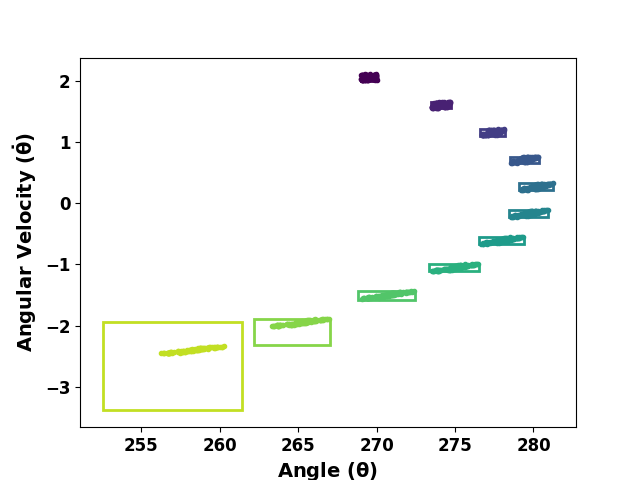}
    \caption{Pendulum}
\end{subfigure}%
\begin{subfigure}{0.33\linewidth}
    \centering
    \includegraphics[width=\linewidth]{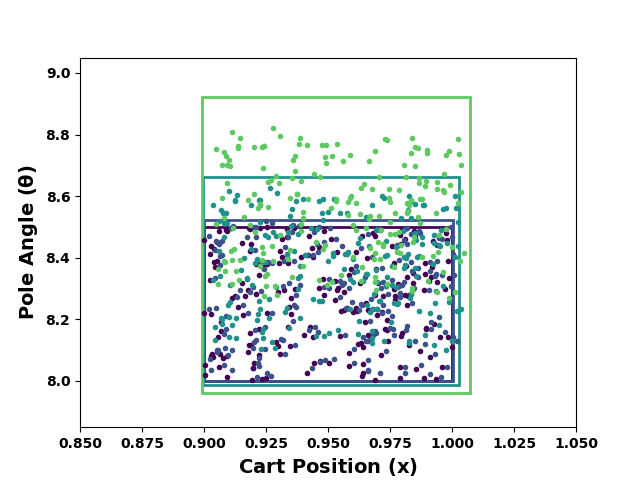}
    \caption{CartPole}
\end{subfigure}%
\begin{subfigure}{0.33\linewidth}
    \centering
    \includegraphics[width=\linewidth]{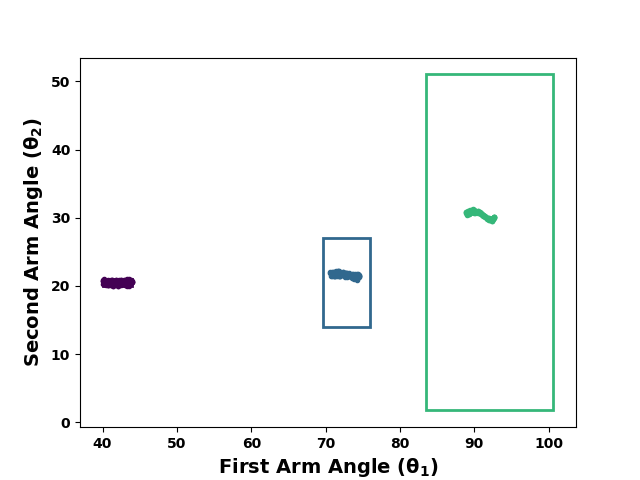}
    \caption{Acrobot}
\end{subfigure}
\caption{Reachable sets over multiple timesteps for three systems. Our method's reachable sets (bounding boxes) are shown with reachable states sampled from the initial conditions (circles).}
\label{fig:reachable_sets_all}
\end{figure*}

In summary, by leveraging transformation specific bounding techniques, we are able to produce informative outer approximations on the observation process, reducing the verification of a vision-based neural controller to a standard problem of neural network verification under input uncertainty.
The final output of this process is an outer-approximation of the one-step reachable set, $\bar{\mathcal{R}}_1(\mathcal{X}_t)$. To extend this analysis for an arbitrary number of timesteps, the pipeline can be applied iteratively,
\begin{equation}
\bar{\mathcal{R}}_T(\mathcal{X}_t) = \underbrace{\bar{\mathcal{R}}_1\big(\bar{\mathcal{R}}_1(\ldots(\bar{\mathcal{R}}_1(}_{T}\mathcal{X}_t)\ldots\big).
\end{equation}
We also expect that tighter $T$-step bounds could be obtained in one shot, as in~\cite{chen2022one}, or with branch-and-bound, as in~\cite{xiang2020reachable,everett2021reachability,li2021tss,bunel2020branch,shi2025neural}.

\section{Experimental Results}
This section demonstrates that the proposed method can generate meaningful reachable set bounds for vision-based, learned control policies.
In particular, we evaluate the method on CartPole, Pendulum, and Acrobot Gym environments \cite{towers2024gymnasium}.
We analyze the method's computational efficiency and bound tightness for various image sizes, up to 100$\times$100$\times$3.

\begin{figure*}[tbh]
\centering
\begin{subfigure}{0.5\linewidth}
    \centering
    \includegraphics[width=1.0\linewidth]{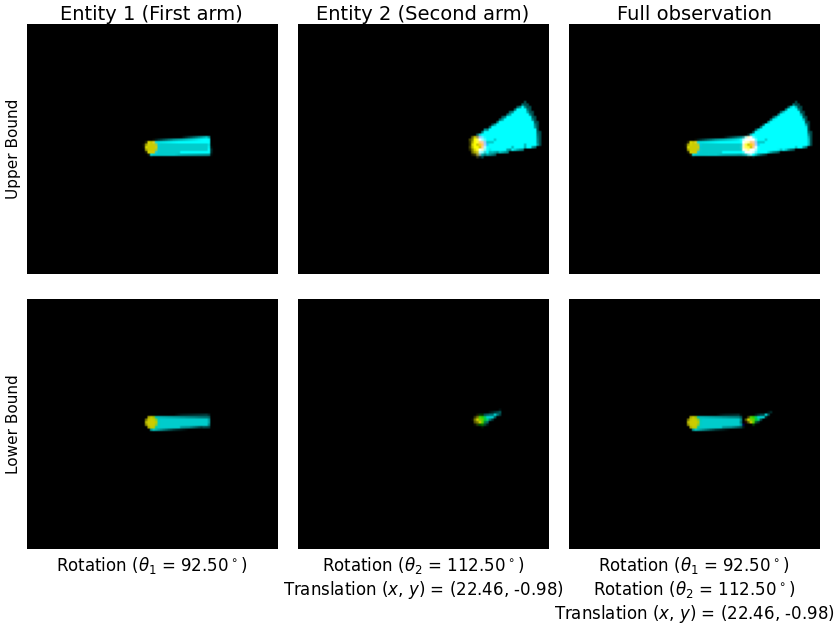}
    \caption{Acrobot}
    \label{fig:obs_set_acro}
\end{subfigure}%
\begin{subfigure}{0.5\linewidth}
    \centering
    \includegraphics[width=1.0\linewidth]{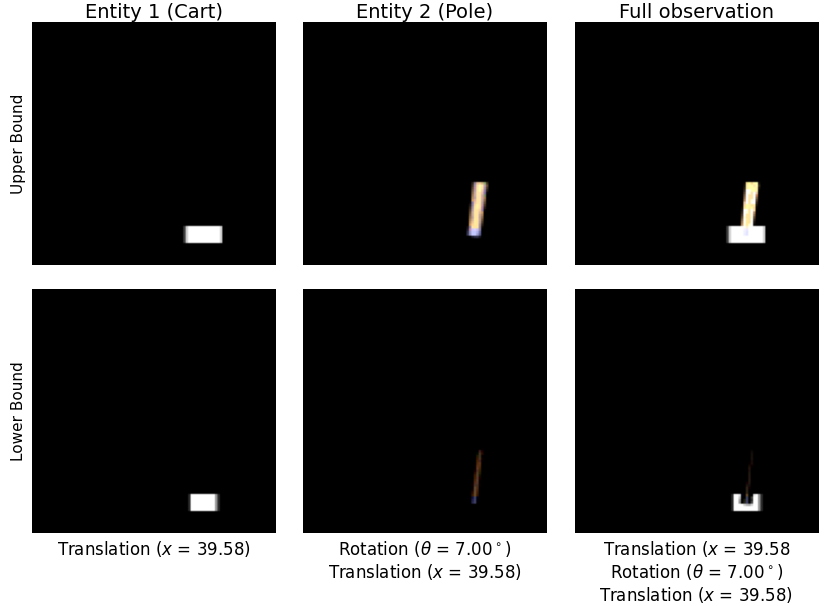}
    \caption{CartPole}
    \label{fig:obs_set_cart}
\end{subfigure}%

\caption{For two systems, a visual representation of image upper (top) and lower (bottom) bounds for each entity (first two columns) and the full blended scene (right column).}
\label{fig:obs_set}
\end{figure*}

\subsection{Experimental Setup}

The experiments use a perception-based controller comprised of a six-layer fully connected neural network. One controller is trained for each environment via Proximal Policy Optimization (PPO) \cite{schulman2017proximal}. Each controller acts on 100$\times$100, 50$\times$50, or 25$\times$25 RGB pixel observations.
All experiments are performed on a machine with an Intel i9 CPU and Nvidia 4090 GPU. 

\subsection{Reachable Sets}
Next, we present empirical results demonstrating the efficacy of our method, beginning with an analysis of the reachable sets $\mathcal{R}_T(\mathcal{X}_t)$.
\cref{fig:reachable_sets_all} visualizes the reachable sets computed using our pipeline on $100 \times 100$ RGB image observations. 
Across all environments, our method produces useful reachable sets several timesteps into the future. 
Notably, for CartPole and Acrobot, prior methods are unable to generate outer reachable sets under geometric transformations even for a single timestep.

As is typical in reachability analysis, the tightness of the bounds degrades for longer time horizons. 
This effect is most prominent in the Acrobot environment, particularly for the angle of the second arm, $\theta_2$. 
This aligns with intuitive expectations, since $\theta_2$ typically exhibits higher variance than $\theta_1$, resulting in visibly wider bounding boxes for $\theta_2$ compared to $\theta_1$.

We also note that the CartPole reachable sets overlap more than other systems, since the goal is to balance the pole upright, the controller tends to make small, alternating movements of the cart, leading to less divergent trajectories and greater overlap in reachable sets over time.

\subsection{Intermediate Analysis: Bounds on Observations}
We next investigate the internal mechanics of our pipeline, focusing specifically on the observation bounds produced by the first stage, denoted $[\underline{Y}, \overline{Y}]$. 
Recall that in this stage, we decompose the scene into individual entities, 
compute bounds for each entity under its corresponding geometric transformation, and then aggregate (or blend) the individual bounds into a unified bound over the full observation.

The upper and lower bounds for each entity, $[\underline{Y}_i, \overline{Y}_i]$, are defined as pixel-wise linear functions of the geometric transformation parameters. 
For example, for transformation $\mathcal{I}_{\mu_i}$ parameterized by $\mu_i$ and pixel $(l, k)$, the lower bound takes the form $\underline{Y}_{i, l, k} = w^{\ell}_{i, l, k} \mu_i + b^{\ell}_{i, l, k}$.
To visualize the quality of these combined bounds, \cref{fig:obs_set_acro,fig:obs_set_cart}, shows the result of evaluating each linear bound at a specific value of $\mu_i$, yielding concrete values for $\underline{Y}_{i, l, k}$ and $\overline{Y}_{i, l, k}$. Each pixel in this figure can thus be interpreted as representing a possible position of the entity under transformation $\mathcal{I}_i$.

We observe that more static components—such as the first arm in Acrobot and the cart in Cartpole—tend to exhibit narrower upper and lower bounds. 
In contrast, more dynamic elements—such as the second arm in Acrobot and the pole in Cartpole—have visibly wider pixel-wise bounds. This observation aligns with our expectations, as these components are known to exhibit greater variability in behavior.
Crucially, we find that both the upper and lower bounds meaningfully capture the spatial extent of each entity, and remain largely consistent after the blending step. 
This indicates that the blending process does not introduce significant slackness, preserving the tightness of the original bounds.

Since the bounds on future states—produced in the second stage of our pipeline—depend on the pixel-wise observation bounds established here, achieving tight state-wise bounds ultimately requires maintaining tight pixel-wise bounds during the first stage.

\begin{figure}[tbh]
    \centering
    \begin{subfigure}{\linewidth}
        \centering
        \includegraphics[width=\linewidth]{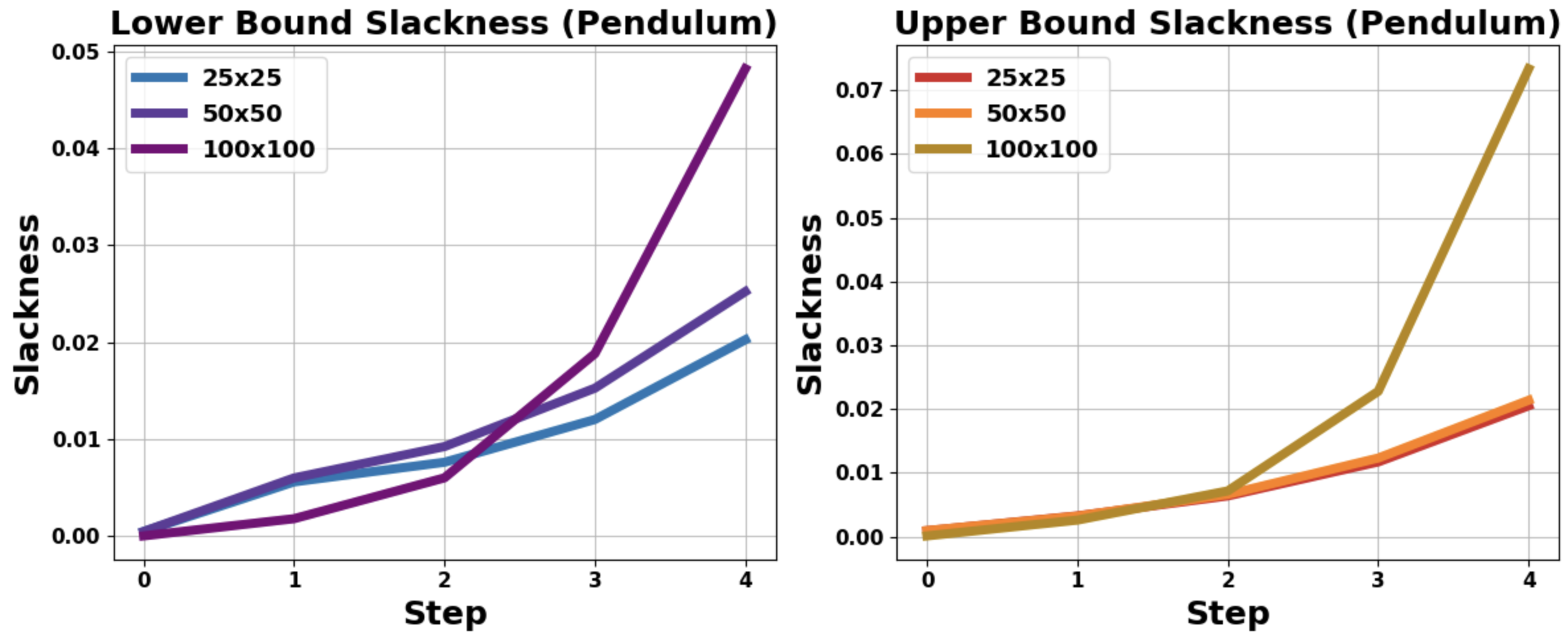}
        \caption{Pendulum}
        
    \end{subfigure}
    \begin{subfigure}{\linewidth}
        \centering
        \includegraphics[width=\linewidth]{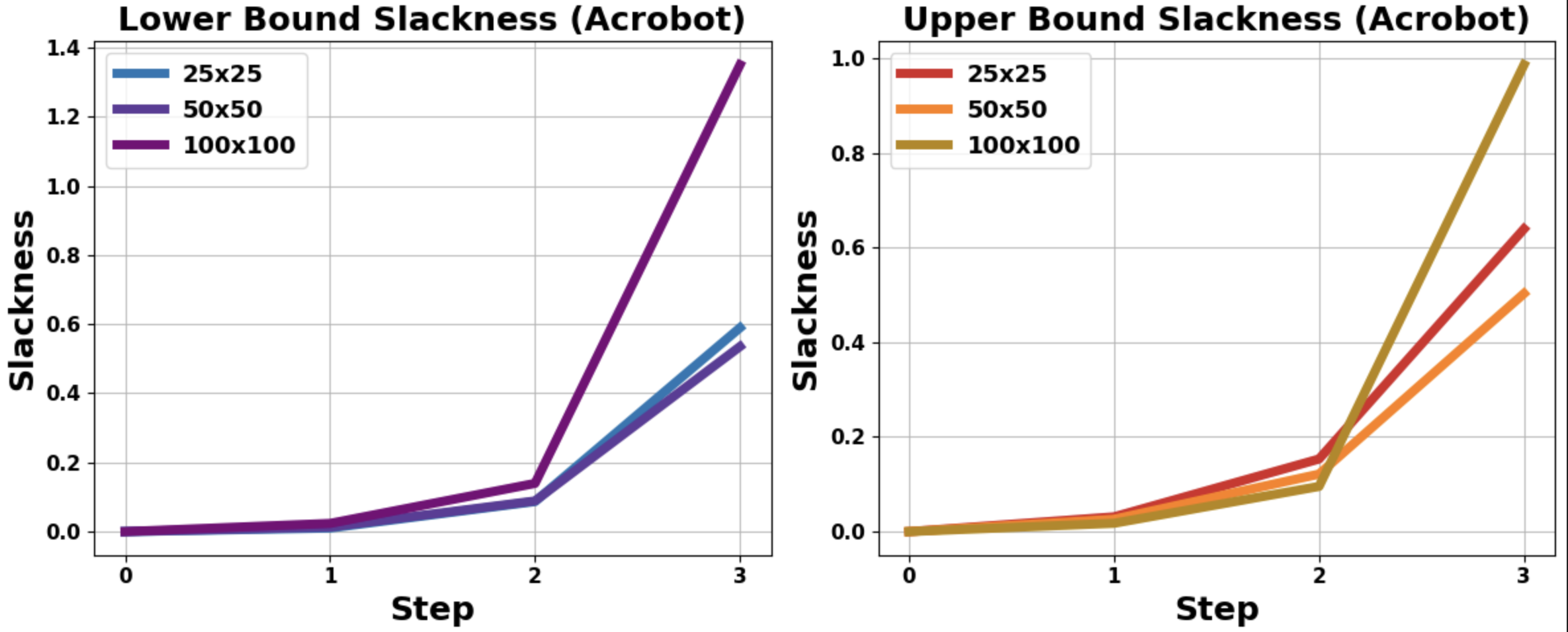}
        \caption{Acrobot}
        
    \end{subfigure}
    \caption{Bound slackness \cref{eq:slack} by image resolutions.}
    \label{fig:slack}
\end{figure}

\subsection{Slackness of Bounds}
While the proposed method computes sound \textit{bounds} on the system's true (intractable to calculate) reachable sets, it is important for these bounds to be tight, otherwise the resulting safety certificates will not be useful.
Therefore, we now analyze the slackness of our reachable sets over multiple timesteps, using sampled states as a surrogate for knowing the true reachable sets.
We define slackness as the minimum distance between any sampled state at timestep $t$ and the corresponding bound. 
Specifically, for a state of dimension $n_x$ and reachable set bounds $\underline{\mathcal{R}}_t(\mathcal{X}_0)$ and $\overline{\mathcal{R}}_t(\mathcal{X}_0)$, we define:
\begin{align}
\text{Upper Slackness} &= \sum_{k=1}^{n_x} \min_{\mathbf{x}_t} \left( \overline{\mathcal{R}}_t^{(k)}(\mathcal{X}_0) - \mathbf{x}_t^{(k)} \right) \label{eq:slack}\\
\text{Lower Slackness} &= \sum_{k=1}^{n_x} \min_{\mathbf{x}_t} \left( \mathbf{x}_t^{(k)} - \underline{\mathcal{R}}_t^{(k)}(\mathcal{X}_0) \right)\nonumber
\end{align}
\begin{figure*}[tbh]
    \centering
    \begin{subfigure}{0.33\linewidth}
        \centering
        \includegraphics[width=\linewidth]{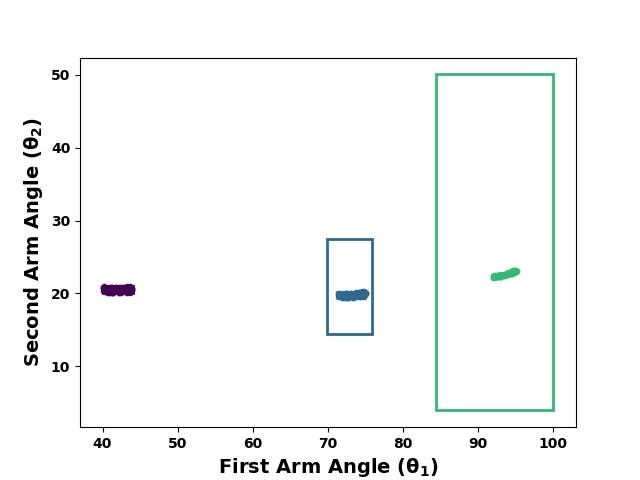}
        \caption{25$\times$25}
    \end{subfigure}%
    \begin{subfigure}{0.33\linewidth}
        \centering
        \includegraphics[width=\linewidth]{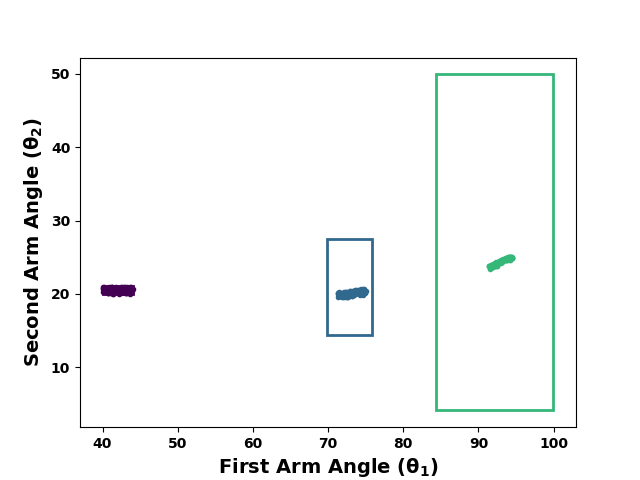}
        \caption{50$\times$50}
    \end{subfigure}%
    \begin{subfigure}{0.33\linewidth}
        \centering
        \includegraphics[width=\linewidth]{plots/main/boundingBoxes/acro_reach.png}
        \caption{100$\times$100}
    \end{subfigure}%
    \caption{Reachable sets on Acrobot for varying image sizes (and thus different control policies).}
    \label{fig:diff_size_reach_set}
\end{figure*}
\cref{fig:slack} presents the slackness values for upper and lower bounds for each environment.
Results are shown for image resolutions of $25 \times 25$, $50 \times 50$, and $100 \times 100$.

As expected, we observe that slackness increases with the prediction horizon in all settings, reflecting the inherent uncertainty growth in long-term predictions. 
We also note that higher-resolution observations tend to yield looser bounds. 
In particular, for Acrobot, the controller trained on $100 \times 100$ images shows a more rapid increase in slackness compared to controllers trained on lower-resolution inputs.
This phenomenon may stem from multiple factors. 
Higher-resolution images may be inherently more difficult to bound. Alternatively, it may reflect behavioral differences in the learned policies—controllers trained on higher-resolution inputs may have adopted more chaotic or less predictable strategies, making them more difficult to tightly bound.

\subsection{Varying Image Size}
Similar to the above, we now examine reachable sets for for varying image sizes of $25\times25$, $50\times50$ and $100\times100$.
\cref{fig:diff_size_reach_set} shows the reachable sets starting from the same initial conditions. 
We again observe that for $100\times100$ images, the produced bounds appear to be slightly less tight than those for $25\times25$ and $50\times50$.
Since each resolution requires its own control policy, varying dynamics in the bounding region can arise from differences in learned control polices.

\subsection{Computational Performance}
Next we investigate the runtime required to produce bounds on observations for each of our three environments.
In \cref{tab:runtime}, the time required to compute bounds over the observation process outlined in \cref{eq:alpha_blend} are shown. 
Note that the time required to compute bounds for each time step is partially dependent on the sampling-based optimization procedure within DeepG; as the width of the interval increases, so to does the runtime for bounding the maximum violation.
As improvements are made on these types of optimizations, the efficiency of our method will also improve.

\begin{table}[tbh]
\centering
\caption{Computational Runtime for Verification of $o(\mathbf{x}_t)$ measured in seconds.}
\begin{tabular}{|c|c|c|c|c|}
\hline
Env & $10 \times 10$ & $25 \times 25$ & $50 \times 50$ & $100 \times 100$\\
\hline
Pendulum  & 0.12 & 0.36  & 1.13 & 4.32 \\
CartPole & 2.48 & 4.83  & 5.96 & 17.28 \\
Acrobot & 2.34 & 4.92  & 6.03 & 26.20 \\
\hline
\end{tabular}
\label{tab:runtime}
\end{table}

\section{Conclusion}
This paper presented a novel verification framework for perception-based controllers operating under realistic geometric perturbations. 
We perform verification using a two-stage pipeline; first, we derive pixel-wise linear bounds on the controller's input under a given set of geometric transformations which act on entities within the full scene, and second, we propagate these bounds through the controller and system dynamics, ultimately yielding bounds on reachable future states. 
By leveraging a state-to-observation mapping, our framework enables the use/adaptation of several state-based reachability techniques even when the controller itself operates entirely on perception.

This work marks a step forward in the verification of closed-loop systems with high-dimensional visual inputs, offering verification for perception-driven controllers under semantically meaningful perturbations (i.e., geometric transformations). 
Our framework is general and can be adapted to different perception models, dynamics, and certification objectives beyond those presented in the paper.
The method provides a tractable and principled pipeline for verification in settings where controllers act on perception, and existing $L^p$-based methods fall short.

\section{Limitations}
While our method offers a step forward in the verification of perception-based controllers, it is not without limitations.

First, although our approach certifies controllers that operate without full access to the state, the verification process itself is still state-dependent. 
That is, our method assumes access to an underlying state and dynamics model in order to propagate the state forward given the actions of the controller. 
We also require that visual scenes can be meaningfully decomposed into individual entities.
For these reasons our approach is well-suited to simulation-based training settings where the underlying state is available (even though the controller does not use the state) but would require further sim-to-real guarantees for real-world scenarios. 
Bridging this gap between simulation-time state access and deployment-time state uncertainty remains a key challenge across all works investigating perception-based controllers.

Second, the worst-case nature of formal verification is often inherently conservative.
This can result in overly pessimistic assessments of controller actions and environmental dynamics, even when the true distributions are \emph{mostly} well behaved. 
As such, while our method is sound, it may limit the operation of a given system due to conservative over-approximation.

\bibliographystyle{IEEEtran}
\bibliography{bib}

\end{document}

%% file: preliminaries.tex
\subsection{Problem Statement}

\begin{figure}[t]
    \centering
    \includegraphics[width=\linewidth]{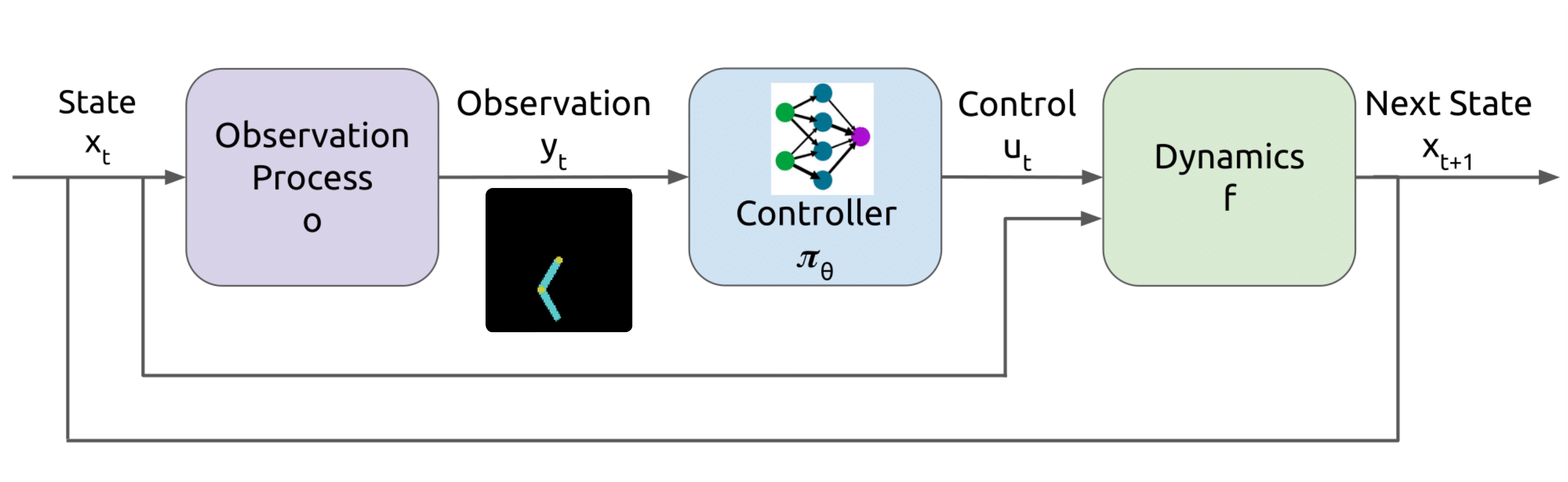}
    \caption{Dynamical System with Vision-Based Control Policy.}
    \label{fig:enter-label}
\end{figure}

For a state space $\mathcal{X} \subseteq \mathbb{R}^{n_x}$, observation space $\mathcal{Y}\subseteq\mathbb{R}^{n_y}$, and control space $\mathcal{U} \subseteq \mathbb{R}^{n_u}$, let there be an observation process $o: \mathcal{X}\to\mathcal{Y}$, a trained control policy $\pi_{\theta}: \mathcal{Y} \to \mathcal{U}$, parameterized by $\theta$, and discrete-time closed-loop dynamics,
\begin{equation}
\mathbf{x}_{t+1} = f(\mathbf{x}_t, \pi_\theta(o(\mathbf{x}_t))), \label{eq:neural_feedback_loop}
\end{equation}
where $\mathbf{x}_t\in\mathcal{X}$ is the system state at time $t$.
Given an initial state set $\mathcal{X}_t \subseteq \mathcal{X}$ and polytope facets $\mathbf{C}\in\mathbb{R}^{c \times n}$, the system's next state $\mathbf{x}_{t+1}$ must be in the 1-step reachable set, $\mathcal{R}_1(\mathcal{X}_t)$, and its  outer bound, $\bar{\mathcal{R}}_1(\mathcal{X}_t)$:
\begin{align}
\mathcal{R}_1(\mathcal{X}_t) &\triangleq \{ \mathbf{x}_{t+1}\ \lvert\ \mathbf{x}_{t+1} = f(\mathbf{x}_{t}, \pi_\theta(o(\mathbf{x}_t))\}\\
&\subseteq \{ \mathbf{x}_{t+1}\ \lvert\ \mathbf{C}\mathbf{x}_{t+1} \leq \mathbf{d}\}\triangleq \bar{\mathcal{R}}_1(\mathcal{X}_t). \label{eq:reachable_set_bounds}
\end{align}
Reachable sets for future timesteps $T$, $\mathcal{R}_T(\mathcal{X}_t)$, can be calculated iteratively.

Calculating $\mathcal{R}_1(\mathcal{X}_t)$ exactly is typically intractable, and thus outer bounds $\bar{\mathcal{R}}_1(\mathcal{X}_t)$ are typically sought. 
Attaining these outer bounds requires bounds on each of the components in \cref{eq:neural_feedback_loop}, namely the dynamics $f$, the control policy, $\pi_{\theta}$, and the observation process $o$.
While existing methods have aimed at bounding $f$ and $\pi_{\theta}$ (as outlined in the next section), it remains a challenge to both bound the observation process as well as to combine bounds on all three components to produce a meaningful outer bound $\bar{\mathcal{R}}_1(\mathcal{X}_t)$. 
To bridge these gaps, we propose a scheme for attaining such outer bounds when observations $\mathbf{y}$ are image based (e.g., $o$ is a render function).

Thus, the goal of our work is twofold:
\begin{enumerate}
    \item We first produce a novel method of bounding the observation process through the use of \emph{geometric transformations}, which are simple functions that inform how changes in state $\mathbf{x}_t$ relate to changes in observation $\mathbf{y}_t$.
    \item With this ability to bound the observation process in terms of simple functions, we are then demonstrate how to integrate these bounds into existing frameworks which give bounds for $f$ and $\pi_{\theta}$, thus producing meaningful outer bounds $\bar{\mathcal{R}}_1(\mathcal{X}_t)$.
\end{enumerate}
Our full pipeline is outlined in greater detail in \cref{sec:method}.

\subsection{Related Work}
\subsubsection{Neural Network Verification}
A large body of work has focused on verifying neural networks under norm-bounded input perturbations, typically $L^p$-balls~\cite{zhang2018efficient, weng2018towards, xu2020automatic, raghunathan2018semidefinite, tjeng2017evaluating, katz2019marabou, katz2017reluplex, jia2021verifying,vincent2021reachable,IBP19,DeepPoly19}.
These techniques aim to provide formal guarantees about a model's output remaining stable despite small input changes.
This paper will leverage Linear Relaxation-based Perturbation Analysis (LiRPA), with details in~\cite{zhang2018efficient}.

\begin{theorem}[Linear Relaxation-based Perturbation Analysis (LiRPA)~\cite{zhang2018efficient}]\label{thm:lirpa}
Given a function $g: \mathcal{X}\subseteq\mathbb{R}^n\to\mathcal{Y}\subseteq\mathbb{R}^m$, input set $\mathcal{X}'\subseteq\mathcal{X}$,
and polytope facets $\mathbf{C}\in\mathcal{R}^{c\times m}$, LiRPA calculates $\mathbf{M},\mathbf{n}$, which define element-wise affine bounds on the network output, $\{\mathbf{y}\in\mathcal{Y}\,\lvert\,\mathbf{C}\mathbf{y}\leq\mathbf{M}\mathbf{x}+\mathbf{n}\}$.
These bounds can be \textit{concretized} to obtain $\mathbf{d}\in\mathbb{R}^c$, which defines polytope outer bounds on the image $g(\mathcal{X}')\subseteq\{\mathbf{y}\in\mathcal{Y}\,\lvert\,\mathbf{C}\mathbf{y}\leq\mathbf{d}\}$.
\end{theorem}

In summary, LiRPA computes $\mathbf{d}$ by forming affine bounds on each primitive in the computation graph, $g$, that are guaranteed to hold over the input domain to that primitive, then aggregates all of these affine bounds, then concretizes the bounds from function input to output (in closed-form for $l_p$-ball input domains).
The codebases~\cite{Bunel2023google,Shi2025Verified} support performing LiRPA on computation graphs with a wide range of primitives that appear in neural networks and dynamical systems (e.g., trigonometric functions, affine transformations, ReLU, sigmoid).
Backward CROWN -- which we will refer to simply as CROWN -- is a type of LiRPA.

\subsubsection{State-Based Verification} \label{sec: nn-ver}
These and similar verification tools have been applied in the context of state-based verification of closed-loop dynamical systems controlled by neural networks (i.e., $o$ is the identity mapping in \cref{eq:neural_feedback_loop}). Techniques have been proposed to propagate reachable sets in time, using symbolic or relaxation-based methods~\cite{sidrane2022overt,chen2023one,everett2021reachability,julian2019reachability,hu2020reach,wang2023polar,ivanov2019verisig,vincent2021reachable,dutta2019reachability, huang2019reachnn, fan2020reachnn, xiang2020reachable,bak2022closed}.

\subsubsection{Image-based Verification}
In a similar vein, image-based verification analyzes policies that operate on visual observations.
Producing exact guarantees in image-based verification is a challenging task, motivating many works to seek either probabilistic or approximate guarantees. Probabilistic methods work by producing bounds or guarantees that hold with high probability \cite{veer2021probably,chou2022safe,watson2025scenario,hsieh2022verifying}, methods in this area typically work via sampling procedure through which guarantees hold with higher probability as more computational time is leverage. On the other hand approximate approaches are designed to work well in practice \cite{chakraborty2024system,chakraborty2024enhancing,chakraborty2023discovering}.

Some methods have also made use of generative models GANs~\cite{GAN22,wu2023toward}, VAEs~\cite{sevin25,waite2023data},  NeRFs~\cite{tong2023enforcing}, primarily to construct low-dimensional latent spaces of plausible observations, allowing the application of state-based verification tools. 
However, these method may not produce true outer approximations as observations or other inputs are generated by learned models.

In some instances formal guarantees are possible, in particular techniques like NNLander-VeriF~\cite{santa2022nnlander} derive formal image constraints directly from a continuous range of rendering parameters (e.g., camera poses), allowing rigorous uncertainty propagation through a visual input pipeline, but only through simple image transformations. 
However, this exactness comes at the cost of only allowing for a single type of modification to the perception (rotation composed with translation).
Differing from this line of work, we investigate verification under geometric transformations in which multiple geometric transformations can effect multiple entities within a given perception, enhancing the practical application of our method.

\subsubsection{Geometric Transformations}
To move beyond norm-bounded robustness, several methods have been proposed to certify neural networks under geometric input transformations, which ensure that transformed images can be represented on a discrete grid with variations in both geometry and appearance.

\begin{definition}[Geometric Transformation]
A \emph{geometric transformation} is a composition of three operations applied to an image domain: a spatial transformation $\mathcal{T}_\mu$, an interpolation scheme $I$, and an affine transformation in pixel intensity $\mathcal{P}_{\psi,\varphi}$.
Given a transformation parameter $\mu$, the spatial component $\mathcal{T}_\mu$ maps image coordinates $(l, k)$ in the transformed image back to continuous-valued coordinates $(\xi_l, \xi_k)$ in the original image domain. The interpolation function $I$, typically bilinear, is then used to obtain the pixel value $p(\xi)$ at these continuous coordinates. Finally, brightness and contrast changes are modeled by the affine mapping $\mathcal{P}_{\psi,\varphi}(p) = \psi p + \varphi$.

Together, these three steps define the full transformation operator $\mathcal{I}_{\psi, \varphi, \mu}$ applied to a pixel at coordinate $(l, k)$, as:
\[
\mathcal{I}_{\psi, \varphi, \mu}(l, k) := \mathcal{P}_{\psi, \varphi} \circ I \circ \mathcal{T}_\mu^{-1}(l, k).
\]
\end{definition}

Several works have extended neural network verification to handle geometric input transformations. Semantify-NN \cite{Semantify20} models geometric transformations—including rotation, translation, and color shift—by introducing symbolic transformation parameters into the network, allowing reuse of norm-bounded verification tools. However, performing bound propagation on image-to-image transformations often results in loose or uninformative bounds. DeepG \cite{DeepG19}, by contrast, computes tight, element-wise affine bounds for transformed pixels by sampling the transformation space and solving a refinement loop to guarantee soundness over a specified parameter range. This yields significantly tighter bounds than interval-based methods.

\begin{theorem}[Pixel-wise Linear Bounds from Geometric Transformations (DeepG) \cite{DeepG19}]\label{thm:deepg}
Given an image $\mathbf{y} \in \mathbb{R}^{H \times W \times C}$, a geometric transformation $\mathcal{I}_{\psi, \varphi, \mu}$ parameterized by $(\psi, \varphi, \mu) \in \mathcal{K} \subseteq \mathbb{R}^d$, and a pixel index $(l,k,c)$, DeepG computes element-wise affine bounds
\[
\underline{p}_{lkc}(\kappa) = w^{\ell}_{lkc}{}^\top \kappa + b^{\ell}_{lkc}, \quad \overline{p}_{lkc}(\kappa) = w^{u}_{lkc}{}^\top \kappa + b^{u}_{lkc}
\]
such that for all $\kappa \in \mathcal{K}$,
\[
\underline{p}_{lkc}(\kappa) \leq \mathcal{I}_{\psi, \varphi, \mu}(x)_{lkc} \leq \overline{p}_{lkc}(\kappa).
\]
\end{theorem}

While these methods consider geometric transformations, they have only been applied to open-loop models (e.g., image classification) and require one geometric transformation to be applied to the entire image.
However, in real scenes, different objects may be subject to different types of movement (e.g., two arms rotating in different directions). To bridge these gaps, we propose a method that handles a multitude of transformations applied to different objects in the scene and is designed for verification of closed-loop systems.